%% file: main.tex
\documentclass[10pt,twocolumn,letterpaper]{article}

\usepackage{cvpr}      % To produce the REVIEW version
\usepackage{xcolor} % add this in your preamble
\usepackage{afterpage}
\usepackage[accsupp]{axessibility} 

\input{preamble}
\definecolor{cvprblue}{rgb}{0.21,0.49,0.74}
\usepackage[pagebackref,breaklinks,colorlinks,allcolors=cvprblue]{hyperref}

\def\paperID{14262} % *** Enter the Paper ID here
\def\confName{CVPR}
\def\confYear{2026}

\newcommand{\name}{OVRCOAT} % [MK] -- I'm just testing an idea...
\newcommand{\cOne}{COAT}
\newcommand{\cTwo}{OVR}% [MK] -- I'm just testing an idea...

\title{Mitigating Objectness Bias and Region-to-Text Misalignment for Open-Vocabulary Panoptic Segmentation}

\author{
    \begin{tabular}{ccc}
        Nikolay Kormushev$^{1,2}$ & Josip Šarić$^{1,3}$ & Matej Kristan$^{1}$ \\
        \multicolumn{3}{c}{\footnotesize} \\ % Switch to smaller size for the rest
        \small $^{1}$University of Ljubljana & \small $^{2}$ETH Zurich & \small $^{3}$University of Zagreb \\ [-0.8ex]
        \small Faculty of Comp. and Inf. Science & \small Dept. of Computer Science & \small Faculty of Elec. Eng. and Computing
    \end{tabular}
}
\begin{document}
\maketitle

\input{sec/0_abstract}    
\input{sec/1_intro}
\input{sec/2_related_work}
\input{sec/3_method}

\input{sec/4_results}
\input{sec/5_conclusion}
\input{sec/6_acknowledgements}

{
    \small
    \bibliographystyle{ieeenat_fullname}
    \bibliography{main}
}

\input{sec/7_supplementary}


\end{document}

%% file: preamble.tex
\usepackage{svg}
\usepackage{float} % in your preamble

\usepackage{pifont}

\newcommand{\xmark}{\ding{55}}  % ✗

\definecolor{gold}{HTML}{BD820B}%{EEAD0E}
\definecolor{silver}{HTML}{909090}%{C0C0C0}
\definecolor{bronze}{HTML}{9A5F26}%{CD7F32}

%% file: sec/0_abstract.tex
\begin{abstract}
Open-vocabulary panoptic segmentation remains hindered by two coupled issues: (i) mask selection bias, where objectness heads trained on closed vocabularies suppress masks of categories not observed in training, and (ii) limited regional understanding in vision–language models such as CLIP, which were optimized for global image classification rather than localized segmentation. We introduce OVRCOAT, a simple, modular framework that tackles both. First, a CLIP-conditioned objectness adjustment (COAT) updates background/foreground probabilities, preserving high-quality masks for out-of-vocabulary objects. Second, an open-vocabulary mask-to-text refinement (OVR) strengthens CLIP’s region-level alignment to improve classification of both seen and unseen classes with markedly lower memory cost than prior fine-tuning schemes. The two components combine to jointly improve objectness estimation and mask recognition, yielding consistent panoptic gains. Despite its simplicity, OVRCOAT sets a new state of the art on ADE20K (+5.5\% PQ) and delivers clear gains on Mapillary Vistas and Cityscapes (+7.1\% and +3\% PQ, respectively). The code is available~\href{https://github.com/nickormushev/OVRCOAT}{here}.
% Understanding complex visual scenes requires models capable of segmenting and recognising not only known categories but also previously unseen objects. Open-vocabulary panoptic segmentation addresses this challenge by unifying instance-level recognition and dense semantic labelling for arbitrary classes. Existing methods such as FC-CLIP and MAFT+ are limited in two ways: the CLIP image encoder is either frozen or only marginally improved during fine-tuning, and background (void) regions are often misclassified, reducing mask quality.
% We propose \name{}, a framework that applies a simple yet effective fine-tuning strategy of the CLIP image encoder for panoptic segmentation, streamlining prior more complex approaches. In addition, \name{} introduces a novel CLIP-based void probability update to guide mask selection, leveraging knowledge learnt during CLIP's large-scale pre-training to mitigate dataset biases and improve foreground-background separation. 
%
%Extensive experiments show that \name{} consistently outperforms prior methods across diverse datasets. On ADE20K, it achieves 28.6 PQ, 15.5 AP, and 34.3 mIoU, improving +1.5 PQ over the previous state-of-the-art MAFT+. Cross-domain evaluation on Mapillary Vistas and Cityscapes yields +1.3 PQ gains, demonstrating strong generalisability. \name{} thus establishes a new state-of-the-art for open-vocabulary panoptic segmentation, providing a robust and generalisable framework for complex scene understanding.
\end{abstract}

%% file: sec/1_intro.tex
\section{Introduction}

Panoptic segmentation provides a semantic visual scene decomposition by assigning a class-label to every pixel, while distinguishing individual object instances (things) from background regions (stuff)~\cite{kirillov2019panoptic}. It thus unifies semantic and instance segmentation, combining class-level labelling with object-level differentiation, 
and delivers a comprehensive representation of objects and their surroundings.

\begin{figure}[t!]
\centering
\includegraphics[width=\linewidth]{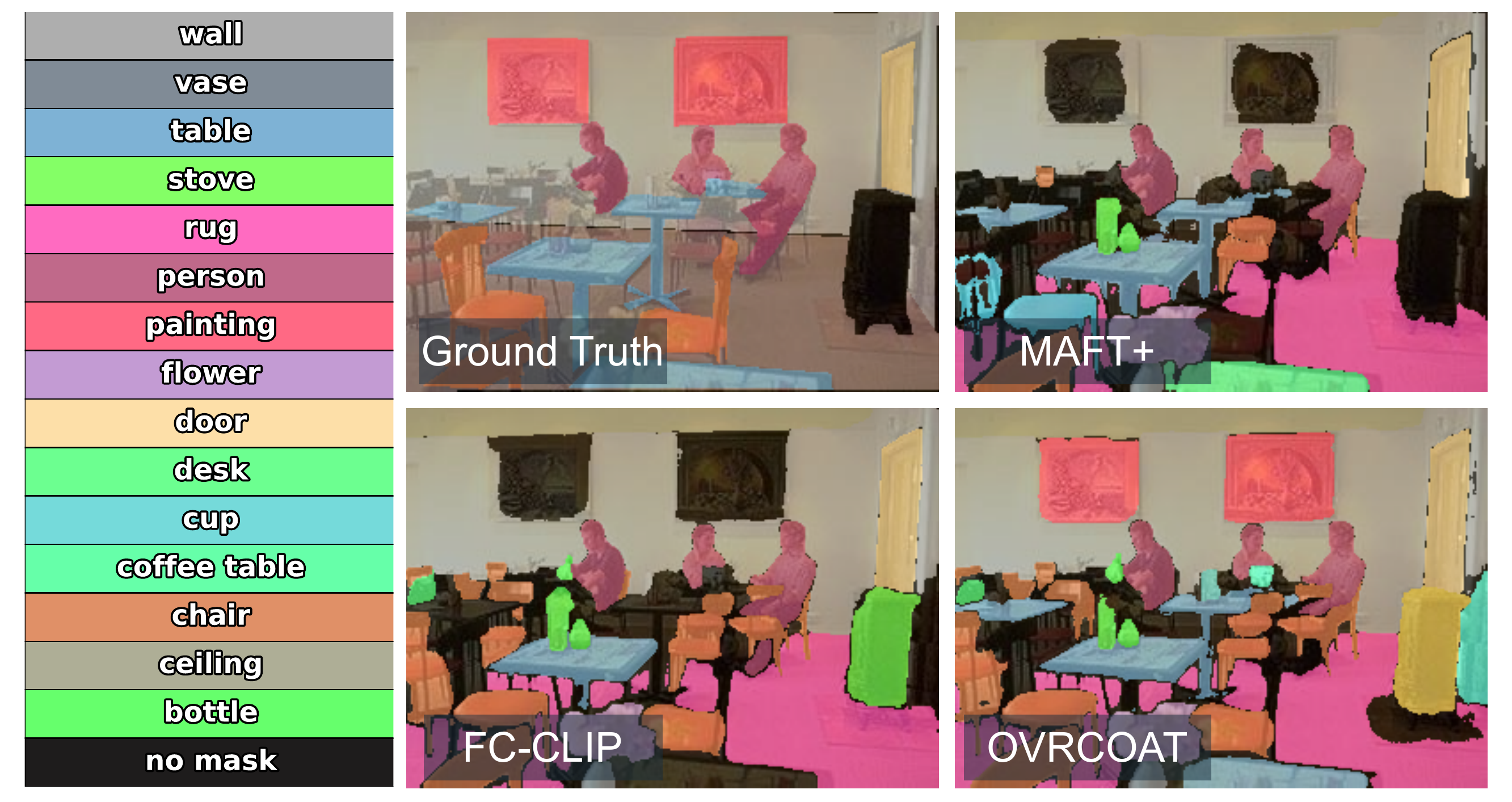}
\caption{
State-of-the-art (e.g., MAFT+~\cite{jiao2024collaborative}, \mbox{FC-CLIP~\cite{yu2023convolutions}})
incorrectly discard detections on categories not present at training time (eg., paintings), leading to failed detection. \name{} correctly classifies these instances by leveraging a CLIP-based update to background probabilities, improving performance on unseen classes. 
}
\label{fig:intro_visual}
\vspace{-1em}
\end{figure}

Conventional panoptic segmentation models~\cite{carion2020end, cheng2021per, cheng2022masked, wang2021max, li2022panoptic, yu2022cmt, jain2023oneformer, li2023mask, sun2023remax} are constrained by a fixed label set (vocabulary) of training data, thus limited only to the categories observed in training. 
This closed-vocabulary restriction poses a major obstacle for real-world deployment~\cite{zhao2017open, yu2023convolutions, xu2023open, jiao2024collaborative, ding2022open}, where visual diversity exceeds the scope of curated datasets, motivating the development of generalisable segmentation paradigms.

One such paradigm is open-vocabulary panoptic segmentation, which leverages vision–language models (VLMs), for example,  CLIP~\cite{radford2021learning} and ALIGN~\cite{jia2021scaling}. 
These methods overcome the aforementioned limitations by first generating individual mask proposals, removing those with low objectness score, and classifying the rest by a VLM-based model. However, this introduces a mask selection bias, since the module for estimating the objectness score is trained on limited data.
A recent study~\cite{saric2025what} noted that in practice, networks often provide very low scores on unlabelled or out-of-training-vocabulary masks, thus classifying them to the background.
\Cref{fig:intro_visual} visualises this bias, where state-of-the-art methods~\cite{yu2023convolutions,jiao2024collaborative} trained on datasets that did not include a \textit{painting} class, incorrectly discard the corresponding masks.

Another challenge is that VLMs, such as CLIP~\cite{radford2021learning} and ALIGN~\cite{ jia2021scaling}, are trained for global image classification rather than image regions localized by segmentation. While early approaches~\cite{xu2023open, yu2023convolutions} relied on frozen CLIP backbones, and proposed work-arounds, recent works such as  
OVSeg~\cite{liang2023open} and MAFT+~\cite{jiao2024collaborative}
propose involved fine-tuning strategies and achieve excellent open vocabulary semantic segmentation. However, in panoptic segmentation, only marginal improvements are observed on datasets like ADE20K~\cite{zhou2017scene} and notable drops on others.

To address the aforementioned challenges, we propose \name{} a novel framework for open-vocabulary panoptic segmentation. The framework is built upon two key components. A CLIP-conditioned objectness adjustment (\cOne{}) leverages CLIP~\cite{radford2021learning} to refine mask selection by distinguishing foreground from background regions, effectively mitigating the mask selection bias present in previous models and improving the overall robustness. 
Open vocabulary mask-to-text feature refinement (\cTwo) introduces an objective that enhances CLIP’s regional understanding and improves mask classification accuracy across both seen and unseen classes, while being considerably more memory-efficient than existing approaches. 
When combined, these two components form \name{}, which jointly improves objectness estimation and mask classification, leading to improved panoptic generalization.
Although fine-tuned specifically for panoptic segmentation, \name{} retains strong generalization across other segmentation tasks. In summary, our main contributions are:
\begin{itemize}
\item CLIP-conditioned objectness adjustment (\cOne) 
mitigates the mask selection bias and improves previously unseen objects detection.
\item Open vocabulary mask-to-text refinement (\cTwo)  
enhances CLIP’s localization and classification capabilities. 
\end{itemize}
Compared to the current state-of-the-art~\cite{jiao2024collaborative}, \name{} proposes a substantially simpler architecture, which brings several benefits. It reduces the memory requirements, future extensions are simpler, and the proposed modules are easily deployable in related methods.
\name{} sets a new state-of-the-art on ADE20K~\cite{zhou2017scene}, outperforming prior open-vocabulary panoptic works by $\mathbf{+5.5\%}$ PQ, and delivers $\mathbf{+7.1\%}$ PQ gains on Mapillary Vistas~\cite{neuhold2017mapillary} and $\mathbf{3\%}$ PQ on Cityscapes~\cite{cordts2016cityscapes}, underscoring its generalisation across different domains.

%% file: sec/2_related_work.tex
 \section{Related Work}

\begin{figure*}[ht!]
  \centering % Centers the figure (and its contents) horizontally
  \includegraphics[width=19cm]{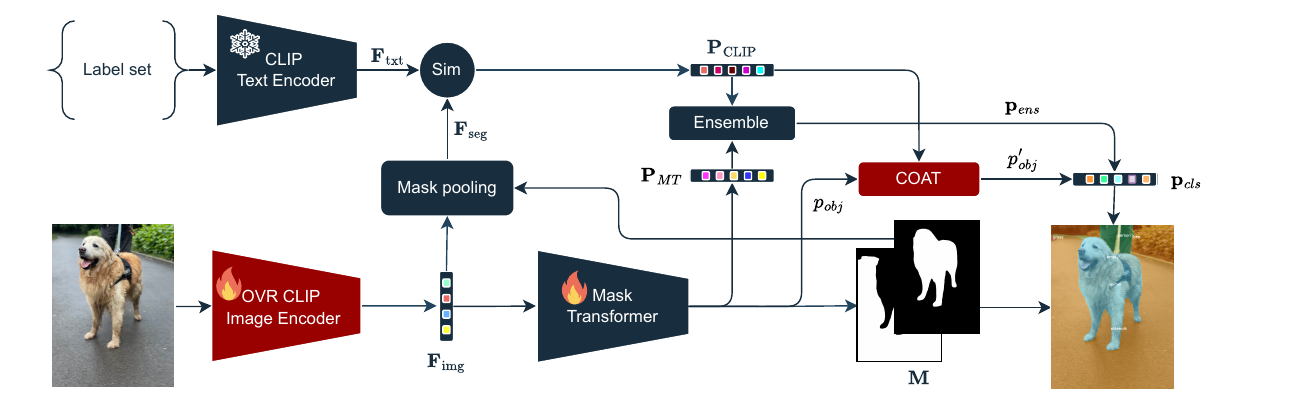} % Use 'width=' for explicit sizing
  \caption{
  OVRCOAT extracts mask-to-text aligned features by CLIP image encoder and mask transformer, adjusted by a novel 
  open-vocabulary mask-to-text refinement finetuning protocol (OVR). A new CLIP-conditioned objectness adjustment (COAT) 
  module reduces the objectness bias in the mask transformer, improving detection of vocabulary instances not observed in training. Our main contributed blocks are denoted by red.
  %\textbf{Overview of the \name{} Open-Vocabulary Panoptic Segmentation Inference Pipeline.} The architecture extends the FC-CLIP framework (components in dark blue, \cite{yu2023convolutions}) by integrating our novel contributions (components in red). The pipeline generates class-agnostic masks and preliminary scores, then obtains CLIP-based classification probabilities by comparing segment embeddings with text embeddings. Our key innovation, a \textbf{Confidence-Aware Probability Fusion} mechanism (red), integrates these outputs, along with an \textbf{Enhanced Void Prediction} module (red), to produce the final panoptic segmentation and improve overall mask selection and foreground-background differentiation.
  }
  \label{fig:coat_overview}
\end{figure*}
\subsection{Vision-language models}

Vision-language models (VLMs) learn mappings of images and text into a shared embedding space by training on large-scale collections of image-caption pairs. Seminal works such as CLIP~\cite{radford2021learning} and ALIGN~\cite{jia2021scaling} adopt a contrastive objective that jointly optimizes image and text encoders to produce close embeddings for positive pairs while maximizing the distance for negative pairs. Such an objective is primarily retrieval-orientated, but also aligns well with classification tasks. These models demonstrate strong zero-shot performance by matching image features with text embeddings of class names, requiring no additional fine-tuning.

However, many downstream tasks, such as semantic and instance segmentation, require pixel-level recognition. Yet the contrastive objective enforces only global image-text alignment, resulting in representations with limited localization capabilities, which are crucial for dense prediction~\cite{rao2022denseclip}. As a result, the naive application of VLMs to pixel-level tasks yields poor segmentation performance.

Various approaches have been considered to address these limitations. Some studies introduce additional pre-training objectives such as masked image reconstruction~\cite{luo2023segclip, tschannen2025siglip}, local-to-global view consistency~\cite{tschannen2025siglip}, and region-to-text alignment~\cite{xu2022groupvit}. In parallel, other studies repurpose pre-trained VLMs for dense prediction through post-training adaptation without relying on dense labels. For instance, these methods improve localization by refining attention mechanisms~\cite{zhou2022extract,wang2024sclip} or by leveraging complementary representations from other foundation models~\cite{wysoczanska2024clip,zhang2025corrclip,wang2025declip,lan2024proxyclip}. Despite significant progress, training-free methods still achieve limited performance and most often lack the ability to distinguish individual object instances. To overcome these limitations, training-based open-vocabulary segmentation methods leverage dense labels to improve localization while retaining the generalization ability of vision–language pretraining.

\subsection{Training-based open-vocabulary segmentation}
While dense semantic labels enable direct fine-tuning of the VLM’s visual encoder, it is beneficial to employ a dedicated decoder to overcome the low output resolution and produce high-resolution segmentation maps. A line of work uses per-pixel decoders that operate on image–text cost volumes and learn to refine the VLM's predictions using dense supervision~\cite{cho2024cat,xie2024sed}. While effective for semantic segmentation, extending these approaches to instance or panoptic segmentation remains highly challenging~\cite{martinovic2025dearli}.

Consequently, most methods adopt mask-transformer decoders, which have proven to be effective architectures for all segmentation tasks~\cite{cheng2021per,cheng2022masked}. Early approaches~\cite{ding2022open,liang2023open} employ mask transformers to generate mask proposals that are then cropped and classified using CLIP. However, this strategy requires multiple evaluations of the CLIP backbone, making it computationally expensive. More efficient variants~\cite{ghiasi2022scaling,xu2023open} mitigate this overhead by extracting image features once and reusing them to classify masks through a mask-pooling operator. FC-CLIP~\cite{yu2023convolutions} further streamlines this design by employing a single frozen convolutional CLIP backbone together with a combined mask-transformer and CLIP-based classifier. MAFT+~\cite{jiao2024collaborative} extends this idea by fine-tuning the backbone for mask classification, while preserving the vision–language alignment of CLIP.
Despite these advances, mask-transformer-based methods still face two key challenges: suboptimal classification performance and missed detections of valid segments caused by biases in the training data~\cite{saric2025what}. Our work tackles these challenges directly.

%Our approach addresses both issues by applying a simple fine-tuning strategy to improve classification accuracy and reclassifying void masks to enhance mask recall.

%Despite these advances, mask-transformer-based methods still face two key challenges: suboptimal classification performance and missed detections of valid segments caused by biases in the training data~\cite{saric2025what}. Our approach addresses both issues by applying a simple fine-tuning strategy to improve classification accuracy and reclassifying void masks to enhance mask recall.

%% file: sec/3_method.tex
\section{\name{}}
\label{sec:method}

We propose \name{}, which features a clean architecture (\Cref{fig:coat_overview}), with two complementary components. A CLIP-conditioned objectness adjustment (\cOne{}) leverages CLIP’s classification confidence to debias the objectness score (Section~\ref{sec:obj_debias_method}), while an open-vocabulary mask-to-text refinement (\cTwo{})
adjusts CLIP’s regional understanding and improves mask-level classification accuracy across both seen and unseen categories (Section~\ref{sec:mask_to_text_align}). The following subsections describe these two components in detail.

\subsection{CLIP-Conditioned Objectness Adjustment}
\label{sec:obj_debias_method}

%As noted in the introduction, 
Mask-transformers-based~\cite{cheng2021per,cheng2022masked} panoptic methods propose a number of potential masks per image, remove the ones with low objectness scores and classify the rest into vocabulary-defined classes~\cite{yu2023convolutions,jiao2024collaborative}. The central interaction between vocabulary classes and objectness score occurs at the definition of the probability density function
% over specified classes, defined as
\begin{equation}
\mathbf{p}_\mathrm{cls} =
\bigl[\mathbf{p}_{\mathrm{ens}} \cdot p_{\mathrm{obj}}, 1 - p_{\mathrm{obj}} \bigr],
\label{eq:FC-CLIP_VOID}
\end{equation}
where $\mathbf{p}_{\mathrm{ens}}$ is the standard ensembled per-vocabulary-class probability density~\cite{yu2023convolutions} and $p_\mathrm{obj}$ is the objectness score (usually defined via $p_\mathrm{obj}=1-p_\mathrm{void}$), acting as a gating function -- a low $p_\mathrm{obj}$ leads to early mask rejection. We make an insight that this score is computed by the mask transformer as a (normalized) similarity of the mask-wide feature with a pre-trained void token. %, relative to the similarities of vocabulary tokens.
It is thus substantially biased towards the training-time vocabulary, in particular, 
if test-time vocabulary instances occur in unlabeled regions of training images, the void token takes form that makes their classification as background more probable (i.e., low objectness).

On the other hand, CLIP does not contain the inherit bias due to its large-scale training, and effectively treats all categories as \textit{seen} classes. We thus propose the following CLIP-based test-time vocabulary-guided bias correction. For a considered proposal mask $\mathbf{M}_i \in \{0, 1\}^{H \times W}$, we extract a feature representation $\mathbf{F}_{\mathrm{seg}, i}$ by a global mask pooling, i.e.,
\begin{equation}
    \mathbf{F}_{\mathrm{seg}, i} =
        \frac{
            \sum_{u,v}^{H,W} \mathbf{M}_i(u,v) \cdot \mathbf{F}_{\mathrm{img}}(u,v)
        }{
        \sum_{u,v}^{H,W} \mathbf{M}_i(u,v)
}.
\end{equation}
A probability distribution that the mask contains any of the test-time vocabulary categories is computed by softmax over dot-products between the mask feature and CLIP-encoded vocabulary terms embeddings
$\mathbf{F}_{\mathrm{txt}}\in\mathbb{R}^{N_\textrm{cls} \times E}$, %i.e.,
\begin{equation}
    \mathbf{p}_{\mathrm{CLIP}, i} = \mathrm{softmax}\Big( \mathbf{F}_{\mathrm{seg}, i} \,           {\mathbf{F}_{\mathrm{txt}}^\top} \Big) \in [0,1]^{1\times N_\textrm{cls}}.  
\end{equation}
If the mask feature agrees well with one of the classes, this results in a high corresponding probability. Alternatively, if it does not agree with any in particular, the distribution is driven towards uniform, effectively reducing the maximal probability. We thus estimate the CLIP-based classification certainty as ${p}_\mathrm{cer}=\max_i \mathbf{p}_{\mathrm{CLIP}, i}$.
Thus the objectness score in (\ref{eq:FC-CLIP_VOID}) is adjusted to 
\begin{equation}
    p_\mathrm{obj}' = 1 - (1- \gamma {p}_\mathrm{cer})(1-p_\mathrm{obj}),
\label{eq:clip_update}
\end{equation}
where $p_\mathrm{obj}$ is the initial mask-transformer objectness score and $\gamma$ is the CLIP trust factor. Note that $\mathbf{p}_\mathrm{cer}$ sets the upper bound on the adjusted objectness $p_\mathrm{obj}'$, with higher values tightening the bound. To account for poor locality in CLIP embeddings, leading to noisy similarities, we set the trust factor to $\gamma=0.5$.

\subsection{Mask to Text Alignment}
\label{sec:mask_to_text_align}

\begin{figure}[ht!]
\centering
\includegraphics[width=1\linewidth]{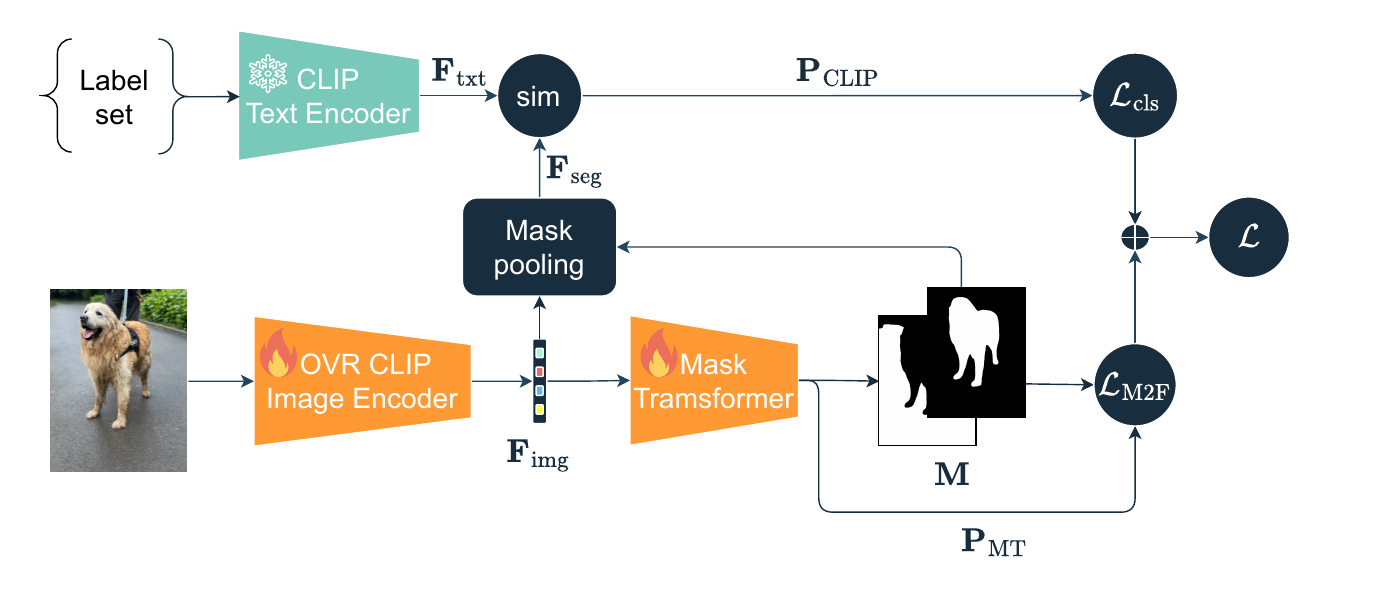}
\caption{
OVR optimizes a combined objective: a mask proposal accuracy $\mathcal{L}_{\mathrm{M2F}}$ and mask classification $\mathcal{L}_{\mathrm{cls}}$ for aligning per-pixel and text embeddings, while preserving vocabulary generalization.
%,improving semantic precision while preserving generalization.
}
\label{fig:training}
\end{figure}

Since CLIP is trained for global image-to-text matching~\cite{radford2021learning}, it's per-pixel features do not align well with object boundaries, leading to inferior mask-level encoding~\cite{shi2024umg}. To address this, we propose open-vocabulary mask-to-text
refinement finetuning protocol (OVR) as shown in~\Cref{fig:training}.

In finetuning, embeddings are extracted by local pooling CLIP image encoder features for each mask proposal generated by the mask transformer. The features are matched to the CLIP embeddings of vocabulary terms and softmaxed, producing the class-wise probability density. Classification is supervised by a cross-entropy loss $\mathcal{L}_{\mathrm{cls}}$
between the obtained class-wise density and the ground truth classification obtained by Hungarian matching~\cite{kuhn1955hungarian} of the proposed masks with ground truth masks. In addition, to supervise mask transformer learning, the classical mask transformer loss~\cite{cheng2022masked} $\mathcal{L}_\mathrm{M2F}$ is applied, leading to the following compounded loss
\begin{equation}
    \begin{aligned}
        \mathcal{L} =  \alpha \, \mathcal{L}_{\mathrm{cls}} + \mathcal{L}_{\mathrm{M2F}}.
    \end{aligned}
    \label{eq:loss_class}
\end{equation}
 
Due to limited panoptic training datasets, care has to be taken in the refinement protocol design. Namely, the limited training-time visual examples and restricted class vocabulary pose overfitting risks, potentially weakening the vision–language alignment generalization.
A two-stage approach is thus proposed. In the first stage, the CLIP image encoder (\Cref{fig:training}) is kept frozen to pre-train the mask generation, without disrupting the CLIP image embedding space. In the second stage, the CLIP image encoder is unfrozen and training continues to jointly tune the embedding-mask alignment and mask proposal generation accuracy. As in~\cite{jiao2024collaborative} the final projection multilayer perceptron and normalization layers in the CLIP image encoder are frozen for restricted training.

%% file: sec/4_results.tex
\begin{table*}[ht!]
\caption{
Analysis on standard datasets. Gray indicates a test set with same vocabulary as used in training. 
%{Panoptic segmentation performance across multiple datasets.} \name{} consistently achieves state-of-the-art results, demonstrating strong cross-domain generalisation. Metrics for Mapillary Vistas and Cityscapes under MAFT+ were computed in this work using the authors’ released code and their panoptic segmentation model, as these results were not reported in the original publication.
}
\label{tab:reclip_results}
\centering
\resizebox{\linewidth}{!}{
\begin{tabular}{l|ccc|ccc|ccc|ccc}
\toprule
Method & \multicolumn{3}{c|}{ADE20k~\cite{zhou2017scene}} & \multicolumn{3}{c|}{Mapillary Vistas~\cite{neuhold2017mapillary}} & \multicolumn{3}{c|}{Cityscapes~\cite{cordts2016cityscapes}} & \multicolumn{3}{c}{\textcolor{gray}{COCO~\cite{lin2014microsoft}}} \\
 & PQ & SQ & RQ & PQ & SQ & RQ & PQ & SQ & RQ & \textcolor{gray}{PQ} & \textcolor{gray}{SQ} & \textcolor{gray}{RQ} \\
\midrule
MaskCLIP~\cite{ding2022open} & 15.1 & 70.5 & 19.2 & - & - & - & - & - & - & \textcolor{gray}{-} & \textcolor{gray}{-} & \textcolor{gray}{-} \\
FreeSeg~\cite{qin2023freeseg} & 16.3 & 71.8 & 21.6 & - & - & - & - & - & - & \textcolor{gray}{-} & \textcolor{gray}{-} & \textcolor{gray}{-} \\
OPSNet~\cite{chen2023open} & 19.0 & 52.4 & 23.0 & - & - & - & 41.5 & 67.5 & 50.0 & \textcolor{gray}{52.4} & \textcolor{gray}{83.5} & \textcolor{gray}{62.1} \\
ODISE~\cite{xu2023open} & 23.4 &\textbf{78.1} & 28.3 & 14.2 & 61.0 & 17.2 & 23.9 & 75.3 & 29.0 & \textcolor{gray}{\textbf{55.4}} & \textcolor{gray}{-} & \textcolor{gray}{-} \\ 
FC-CLIP~\cite{yu2023convolutions} & 26.8 & 71.2 & 32.3 & 18.3  & 56.0  & 23.1 & 44.0 & 75.4 & 53.6 & \textcolor{gray}{54.4} & \textcolor{gray}{-} & \textcolor{gray}{-}\\
MAFT+\textsubscript{pan}~\cite{jiao2024collaborative} & 27.1 &  73.5  & 32.9  & 15.7 & 55.5 & 19.8 & 38.3 & 70.2 & 46.9 & \textcolor{gray}{50.3} & \textcolor{gray}{82.2} & \textcolor{gray}{60.3}   \\
\midrule
\name{} (Ours) &  \textbf{28.6}\textsubscript{+1.5} & 77.3\textsubscript{-0.8} & \textbf{34.7}\textsubscript{+1.8} & \textbf{19.6}\textsubscript{+1.3} & \textbf{65.7}\textsubscript{+4.7} & \textbf{24.8}\textsubscript{+1.7} & \textbf{45.3}\textsubscript{+1.3} & \textbf{78.7}\textsubscript{+3.3} & \textbf{55.6}\textsubscript{+2.0} & \textcolor{gray}{54.6} & \textcolor{gray}{82.9} & \textcolor{gray}{65.1}\\
\bottomrule
\end{tabular}
} \vspace{-1em}
\end{table*}

\section{Experiments}
%\subsection{Implementation \& training  details}

\name{} uses the standard backbone, i.e., ConvNeXt-Large~\cite{radford2021learning, liu2022convnet, yu2023convolutions} from OpenCLIP~\cite{ilharco2021openclip}, pretrained on LAION-2B~\cite{schuhmann2022laion}. The mask transformer architecture follows standard Mask2Former~\cite{cheng2022masked}. We train \name{} using the AdamW optimiser with a weight decay 0.05, using a  learning rate $1\times10^{-4}$ in the first-stage training, and $5\times10^{-5}$ in the second stage. The classification loss $\mathcal{L}_\mathrm{cls}$ weight in (\ref{eq:loss_class}) is set to 
$\alpha=0.1$, while we use the same parameters as in~\cite{cheng2022masked} to train the mask transformer. The trust factor in (\ref{eq:clip_update})  is set to $\gamma = 0.5$.
We employ three NVIDIA A100 GPUs (40~GB each), and a batch size of 9.

In all of our experiments, the COCO~\cite{lin2014microsoft} closed-vocabulary panoptic segmentation dataset is used for training \name{}.
%\name{} is trained COCO~\cite{lin2014microsoft} closed-vocabulary panoptic segmentation dataset for all our experiments. 
We then evaluate it under two setups: open-vocabulary panoptic segmentation (\Cref{sec:ovPS}) and semantic segmentation (\Cref{sec:SemSeg}).

\subsection{Open-Vocabulary Panoptic Segmentation}
\label{sec:ovPS}

\name{} is compared with several state-of-the-art methods, i.e., MaskCLIP~\cite{ding2022open}, FreeSeg~\cite{qin2023freeseg}, OPSNet~\cite{chen2023open}, ODISE~\cite{xu2023open}, FC-CLIP~\cite{yu2023convolutions} and the panoptic version of the recent state-of-the-art MAFT+~\cite{jiao2024collaborative} (MAFT+$_\mathrm{pan}$) on three challenging out-of-vocabulary (OOV) datasets ADE20K~\cite{zhou2017scene}, Cityscapes~\cite{cordts2016cityscapes} and Mapillary Vistas~\cite{neuhold2017mapillary}. Standard performance measures~\cite{kirillov2019panoptic} are used: Panoptic Quality (PQ), Segmentation Quality (SQ) and Recognition Quality (RQ).  

Results are reported in \Cref{tab:reclip_results}. 
\name{} consistently outperforms previous methods across out-of-vocabulary datasets, on average outperforming the state-of-the-art MAFT+ in PQ by approximately $16\%$.
%achieving Panoptic Quality (PQ) improvements %ranging from 1.3 to 1.5 compared to the current state-of-the-art. 
Only on COCO~\cite{lin2014microsoft}, \name{} slightly lags behind ODISE~\cite{xu2023open}, with a 1.4\% drop.  
%This can be attributed to the CLIP-based update, which is primarily designed to improve generalisation to unseen classes, at a cost of small accuracy reduction on categories seen during base model training (we further explore the reasons in \cref{tab:training_void_ablation}). 
This can be attributed to the COAT module, which is designed to improve generalisation by leveraging CLIP~\cite{radford2021learning} for mask selection. While it reduces training biases in unseen classes, replacing the original specialised mask evaluation, which was trained on the COCO dataset, can hinder performance in those seen classes (we verify this empirically in \Cref{sec:performance_unseen}). Nevertheless, despite this effect, \name{} substantially outperforms ODISE on all other datasets, by up to $90\%$ on Cityscapes~\cite{cordts2016cityscapes}.

%We present our main panoptic segmentation results in \cref{tab:reclip_results}. \name{} consistently outperforms previous methods across out-of-vocabulary (OOV) datasets, achieving Panoptic Quality (PQ) improvements ranging from 1.3 to 1.5\% compared to the leading baselines.
%However, we observe that the in-vocabulary performance on COCO~\cite{lin2014microsoft} is slightly lower than that of ODISE~\cite{xu2023open}. This difference is attributable to our CLIP-based void update, which is designed to improve generalisation to unseen classes but may reduce accuracy on categories seen during the base model's training. In support of this analysis, our training-only void variant performs on par with ODISE, as detailed in \cref{tab:training_void_ablation}, confirming the efficacy of the proposed fine-tuning scheme.

\name{} achieves substantial gains in the component metrics. %We improve Segmentation Quality (SQ), 
In particular, on Mapillary Vistas~\cite{neuhold2017mapillary}, a remarkable $18\%$ SQ performance improvement is observed compared to the current state-of-the-art MAFT+~\cite{jiao2024collaborative} and $7.7\%$ compared to ODISE~\cite{xu2023open}. 
%In particular, on Mapillary Vistas~\cite{neuhold2017mapillary}, a remarkable +10.2 SQ points performance improvement is observed compared to the current state-of-the-art MAFT+~\cite{jiao2024collaborative}, and +4.7 compared to ODISE~\cite{xu2023open}. 
An approximate $5.5\%$ RQ improvement compared to top per-dataset performer is observed across all datasets. These results demonstrate the strong cross-domain generalisation of \name{}. The generalisation result is particularly strong in the context of comparatively large MAFT+ PQ degradation outside of ADE20k~\cite{zhou2017scene}, highlighting the challenges of fine-tuning.

%whereas MAFT+ exhibits PQ degradation outside of ADE20k~\cite{zhou2017scene}, highlighting the limitations of its fine-tuning strategy.

%while Recognition Quality (RQ) increases by roughly 2 points across datasets. In conclusion, these results demonstrate the strong cross-domain generalisation of \name{}, whereas MAFT+ exhibits PQ degradation outside of ADE20k~\cite{zhou2017scene}, highlighting the limitations of its fine-tuning strategy.
\subsection{Ablation study}

% \subsubsection{Impact of COAT and Mask-to-text Refinement }
We analyse the individual contributions of the proposed CLIP-conditioned objectness adjustment (\cOne{}) and open-vocabulary mask-to-text refinement (\cTwo{}) to panoptic segmentation performance. The results are reported in \Cref{tab:training_void_ablation}.
%To better understand the individual contributions of our training procedure and the CLIP-based~\cite{radford2021learning} void probability update, we evaluate them separately. The results of this ablation study are presented in $\cref{tab:training_void_ablation}$.

We first isolate the effect of the proposed COAT by applying it to our baseline (FC-CLIP~\cite{yu2023convolutions}). This modification alone consistently improves PQ across all out-of-vocabulary datasets, demonstrating gains ranging from 1.4\% to 3\%.
We next evaluate the contribution of the mask-to-text refinement alone.
The performance drops by $3.6\%$ in PQ compared to original \name{}, but improves by $1\% - 5\%$ compared to the baseline. Our training strategy, applied on the COCO dataset, is effective not only in enhancing performance on its validation set, encouraging CLIP to focus on the dataset’s distribution, but also in preserving a strong generalisation to unseen domains. 
%This verifies that our training strategy is effective not only in enhancing performance on the COCO training set, where fine-tuning encourages CLIP to focus more strongly on the dataset’s distribution and object categories, but also in generalising performance to unseen domains.
Combining COAT and \cTwo{}, producing \name{}, delivers clear advantages over the baseline, with improvements of up to 7\% PQ, validating the complementary contributions of the two components, while incurring a small trade-off in performance on seen classes. Despite this, \name{} still surpasses most state-of-the-art models (\Cref{tab:reclip_results}).

%Next, we validate our training approach by listing a train-only variant of \name{} (without the final void update). This variant consistently improves PQ over FC-CLIP~\cite{yu2023convolutions}, with gains ranging from +0.5 to +1.1 PQ. These results confirm that our training strategy is effective not only in enhancing performance on the COCO training set but also in generalising performance to unseen domains.

%\name{} is the result of combining the improved trained model with the CLIP-based void probability update, achieving our best overall results. Performance improvements range from +0.4 to +1.0 PQ, depending on the dataset. The only exception is the COCO dataset, where performance slightly decreases compared to the train-only variant. This reflects the expected trade-off between achieving stronger cross-domain generalisation and maximising seen-class performance.

\begin{figure*}[ht!]
\centering
\includegraphics[width=\linewidth]{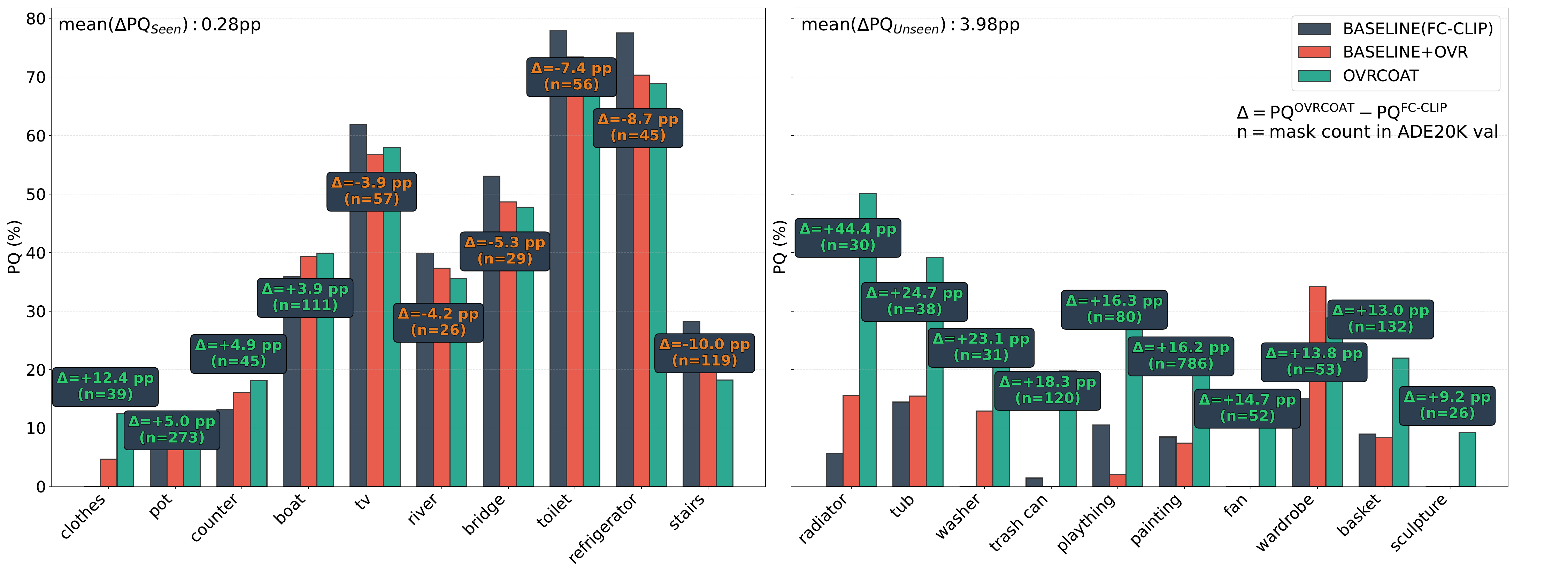}
\caption{
Per-class PQ differences between \name{} and FC-CLIP~\cite{yu2023convolutions} on ADE20k~\cite{zhou2017scene}
for classes with largest PQ differences.
Mean values are reported weighted with ground truth class frequency (weighted), and treating all classes equally (unweighted).
\name{} yields substantial improvements on unseen categories, demonstrating stronger generalisation, with stable performance on seen classes.
%Largest absolute changes in panoptic quality (PQ) for individual classes are shown. \name{} yields substantial improvements on unseen categories, demonstrating stronger generalisation, while performance on seen classes remains largely stable. Mean values are reported both frequency weighted, according to ground-truth class frequencies, and unweighted, treating all classes equally.
}\vspace{-0.5em}
\label{fig:pq_difference}
\end{figure*}
% \begin{table}[ht!]
% \caption{\textbf{Analysis of individual contributions.} Evaluation of the COAT module and Mask to Text Alignment, illustrating their complementary roles in achieving the final \name{} performance.}
% \label{tab:training_void_ablation}
% \centering
% \resizebox{\linewidth}{!}{
% \begin{tabular}{l|cccc}
% \toprule
% Method & ADE20k%~\cite{zhou2017scene} 
% & Mapillary %Vistas~\cite{neuhold2017mapillary} 
% & Cityscapes%~\cite{cordts2016cityscapes} 
% & COCO%~\cite{lin2014microsoft} 
% \\
% \midrule
% FC-CLIP~\cite{yu2023convolutions} & 26.8 & 18.3 & 44.0 & 54.4 \\
% FC-CLIP_\textrm{COAT} & 27.6 & 18.8 & 44.6 &  53.7 \\
% \name{}_\textrm{align only} & 27.6 & 19.2 & 44.5 & \textbf{55.5} \\
% \name{} & \textbf{28.6} & \textbf{19.6} & \textbf{45.3} & 54.6 \\
% \bottomrule
% \end{tabular}
% }
% \end{table}

\begin{table}[ht!]
\caption{
Ablation study of OVRCOAT individual modules.
%\textbf{Analysis of individual contributions.} Evaluation of the COAT module and Mask to Text Alignment, illustrating their complementary roles in achieving the final \name{} performance.
}
\label{tab:training_void_ablation}
\centering
\resizebox{\linewidth}{!}{
\begin{tabular}{cc|cccc}
\toprule
\cOne{} & \cTwo{} & ADE20k%~\cite{zhou2017scene} 
& Mapillary %Vistas~\cite{neuhold2017mapillary} 
& Cityscapes%~\cite{cordts2016cityscapes} 
& COCO%~\cite{lin2014microsoft} 
\\
\midrule
\xmark & \xmark & 26.8 & 18.3 & 44.0 & 54.4 \\
\checkmark & \xmark & 27.6 & 18.8 & 44.6 &  53.7 \\
\xmark & \checkmark & 27.6 & 19.2 & 44.5 & \textbf{55.5} \\
\checkmark & \checkmark & \textbf{28.6} & \textbf{19.6} & \textbf{45.3} & 54.6 \\
\bottomrule
\end{tabular}
}
\end{table}

% \subsubsection{Per-class performance analysis}
\label{sec:performance_unseen}
To more precisely disentangle the improvements introduced by \name{}, we additionally analyze per-class panoptic quality (PQ) on ADE20K. We stratify categories into seen and unseen groups depending on whether they appear in the training set. \Cref{fig:pq_difference} shows per-class PQ differences between the baseline and \name{} for the 10 seen and 10 unseen classes with the largest absolute change, measured as $\lvert \text{PQ}_{\text{FC-CLIP}}(c) - \text{PQ}_\text{\name{}}(c) \rvert$ for each class $c$.
Additionally, we report the baseline enhanced with OVR, allowing us to isolate the specific contribution of \cOne{} beyond this component.

%As shown in , we report the top 10 seen and top 10 unseen classes with the largest absolute changes in PQ. 
For classes seen during training, \name{} exhibits a mixed effect: categories less common in the COCO training set ($\approx$ 5000 masks on average), such as \textit{apparel}, \textit{boat}, and \textit{counter}, show clear gains, while those more frequently observed during training ($\approx$ 14 000 masks on average) experience only a moderate decrease. 
On average, panoptic quality across seen classes is minimally reduced for -0.05~pp, indicating that performance is largely preserved, while a +3.9~pp improvement is observed on unseen classes.

Conversely, unseen classes gain substantially from the combined effect of our mask-to-text refinement training strategy and the \cOne{} module. For example, a dramatic relative improvement of $192\%$ is observed for category \textit{paintings}, which is commonly misclassified as background by the current state-of-the-art~\cite{vsaric2025holds}. Overall, average relative improvement of 25\% is observed across unseen classes.
% These classes achieve a 16.3\% increase in PQ. Across all unseen categories, the average improvement in PQ is approximately 4\%. 
These results clearly highlight that \name{} enhances generalisation to novel classes while simultaneously maintaining competitive accuracy on those seen during training.

\subsubsection{Semantic Segmentation}
\label{sec:SemSeg}

For further insights, we evaluate the impact of our contributions on the semantic segmentation performance.
\Cref{tab:open_vocab_semseg_best} presents the results in terms of standard mean Intersection over Union (mIoU) on the PASCAL Context~\cite{mottaghi_cvpr14} and ADE20K~\cite{zhou2017scene} datasets. The first section presents the results of mask-transformer-based approaches from the literature. The second section 
incrementally presents the impact of our contributions \cTwo{} and \cOne{}.
% Preliminary experiments show that semantic segmentation performance is slightly reduced when applying COAT. We explore the reasons in detail in ~\Cref{sec:train_impact} and report the analysis of \name{} without using the aforementioned module (\name{}$_\textrm{\cTwo~only}$).

We first establish the notion of slight disadvantage of panoptic models compared to semantic segmentation models when evaluated on the semantic segmentation task.
The results in \Cref{tab:open_vocab_semseg_best} show that the current state-of-the-art MAFT+ variant trained specifically for semantic segmentation (MAFT+$_\textrm{sem}$) achieves the highest overall mIoU. It consistently outperforms the variant trained with panoptic labels (MAFT+$_\textrm{pan}$) by 1–9\% in relative terms.
This suggests that the additional demands of panoptic recognition consume part of the model’s capacity, thereby degrading the semantic segmentation performance.

Our open-vocabulary mask-to-text refinement (\cTwo{}) improves semantic segmentation performance over the baseline (FC-CLIP) in 4 out of 5 datasets,
despite originally designed for panoptic segmentation.
On the other hand, further inclusion of \cOne{}
slightly degrades the semantic segmentation performance. This behavior stems from the way mask transformers perform semantic inference. Unlike in panoptic segmentation, no masks are rejected. Instead, all predicted masks contribute to the aggregated class distribution at every pixel. As a result, masks that would otherwise be discarded in panoptic inference due to low objectness, still influence the final per-pixel predictions. Hence, adjusting objectness has little real effect and actually introduces noise due to CLIP’s weaker localization.

% However, \name{} improves the panoptic version MAFT+$_\textrm{pan}$'s mIoU on ADE20k by 0.7\% across all classes and 2.6\% on the 150-class subset. On Pascal Context~\cite{mottaghi_cvpr14}, the results are comparable, ranging from  between -3.5\% and 0.8\% difference.

%However, for their panoptic segmentation model, \name{} surpasses MAFT+ on ADE20k~\cite{zhou2017scene}, achieving +0.1 mIoU across all classes and +0.5 on the 150-class subset, while the results on Pascal Context~\cite{mottaghi_cvpr14} are mixed, ranging from -0.7 to +0.5 PQ difference.

\begin{table}[ht!]
\caption{Open-vocabulary semantic segmentation performance (mIoU) across multiple datasets, with best performance bolded. 
% MAFT+$_\textrm{sem}$ and MAFT+$_\textrm{pan}$ are MAFT+~\cite{jiao2024collaborative} trained on segmentation and panoptic labels, respectively.
}
\label{tab:open_vocab_semseg_best}
\centering
\resizebox{\linewidth}{!}{
\begin{tabular}{l c|ccccc}
\toprule
Model & Backbone & A-847 & A-150 & PC-459 & PC-59 & PAS-20 \\
\midrule
OpenSeg~\cite{jia2021scaling} & ALIGN & 8.8 & 28.6 & 12.2 & 48.2 & 72.2 \\
OVSeg~\cite{liang2023open} & ViT-L & 9.0 & 29.6 & 12.4 & 55.7 & 94.5 \\
SAN~\cite{xu2023side} & ViT-L & 12.4 & 32.1 & 15.7 & 57.7 & 94.6 \\
ODISE~\cite{xu2023open} & ViT-L & 11.1 & 29.9 & 14.5 & 57.3 & - \\
FC-CLIP*~\cite{yu2023convolutions} & ConvNeXt-L & 14.8 & 34.0 & 18.2 & 58.4 & 95.4 \\
MAFT+\textsubscript{sem}~\cite{jiao2024collaborative} & ConvNeXt-L & \textbf{15.1} & \textbf{36.1} & \textbf{21.6} & \textbf{59.4} & \textbf{96.5} \\
MAFT+\textsubscript{pan}~\cite{jiao2024collaborative} & ConvNeXt-L & 14.2 & 33.8 & 19.8 & 58.0 & 95.6 \\
\midrule
\cTwo{} & ConvNeXt-L & 14.3 & 34.3 & 19.1 & 58.5 & 95.5 \\
\name{} & ConvNeXt-L & 14.0 & 33.7 & 18.1 & 55.4 & 94.1 \\
\bottomrule
\end{tabular}
}\vspace{-0.5em}
\end{table}

\subsection{Architecture Design Validation Study}

\subsubsection{Mask Classifier Validation}
Since \name{} fine-tunes the CLIP~\cite{radford2021learning} image encoder on COCO~\cite{lin2014microsoft}, it is natural to investigate whether using only CLIP suffices for mask classification. \Cref{tab:ensemble_ablation} evaluates four different setups: the standard ensemble of mask-transformer and our CLIP\textsubscript{\cTwo{}} predictions~\cite{yu2023convolutions} with and without \cOne{}, and using the fine-tuned CLIP\textsubscript{\cTwo{}} alone, again with and without \cOne{}.
The results show that the ensemble with \cOne{} (as used in \name{}) outperforms all other variants. In particular, leveraging the mask-transformer probabilities as a stronger specialist on seen classes yields around a 14\% relative improvement over using CLIP alone in both configurations. We also observe that \cOne{} consistently boosts performance for both the ensemble and the standalone CLIP-based mask classifier.
% using the ensembling with the mask transformer probabilities (\name{}\textsubscript{\cTwo~only}), using only CLIP predictions (\name{}$_\textrm{no ensemble}$), and using only CLIP predictions while applying the COAT module for the void probability(\name{}$_\textrm{void fusion}$), which biases mask selection toward the masks for which CLIP is more confident. 

%Since we fine-tuned the CLIP~\cite{radford2021learning} image encoder on the COCO dataset~\cite{lin2014microsoft}, it is natural to investigate whether only CLIP suffices for mask classification. To explore this, we compare three setups: using the geometric ensemble, using only CLIP predictions, and using only CLIP predictions but with the void update, which biases mask selection toward those masks in which CLIP is more confident. 

\begin{table}[ht!]
\caption{
Comparison of various mask classification methods. 
%Evaluating the impact of probability fusion on mas classification. \name{}$_\textrm{void fusion}$ denotes applying fusion only to the void probability.
}
\label{tab:ensemble_ablation}
\centering
\begin{tabular}{lc|ccc}
\toprule
Mask Classifier & COAT & PQ & SQ & RQ\\
\midrule
% \name{}\textsubscript{\cTwo~only} & 27.6 & 74.1 & 33.4 \\
% \name{}\textsubscript{no fusion} & 24.2 & 75.9 & 29.2 \\
% \name{}\textsubscript{void fusion} & 24.9 & 78.4 & 30.2 \\
CLIP\textsubscript{OVR} & \xmark & 24.2 & 75.9 & 29.2 \\
CLIP\textsubscript{OVR} & \checkmark & 24.9 & 78.4 & 30.2 \\
Ensemble & \xmark & 27.6 & 74.1 & 33.4 \\
Ensemble & \checkmark & 28.6 & 77.3 & 34.7 \\
\bottomrule
\end{tabular}
\end{table}

\subsubsection{Classification of Perfect Masks}
To further evaluate the effectiveness of our \cTwo{} fine-tuning approach and its impact on seen and unseen classes, we analyse the performance using a segmentation oracle. The oracle generates perfect binary masks, one for each ground truth panoptic segment, replacing those produced by the mask transformer. During inference, we follow the \name{} pipeline, but completely omit the mask transformer, relying solely on the CLIP for classification.

We compare two variants of CLIP~\cite{radford2021learning} using this oracle: the pre-trained OpenCLIP ConvNeXt-Large~\cite{ilharco2021openclip, liu2022convnet}, referred to as CLIP, and our \cTwo{} fine-tuned version (CLIP\textsubscript{\cTwo{}}). Results in \Cref{tab:perfect_masks} show an 11\% improvement on PQ in favour of CLIP\textsubscript{\cTwo{}}. We also observe substantial gains of 9.2\% on seen classes and 13\% on unseen classes, indicating that our fine-tuning not only enhances performance on trained categories, but also generalises effectively. These results indicate that \cTwo{} effectively adapts CLIP to open-vocabulary panoptic segmentation.

\begin{table}[h]
\centering
\caption{Evaluation of \cTwo~ 
using oracle (ground truth) masks.}
\begin{tabular}{l|ccc}
\toprule
Model & PQ & PQ\textsubscript{seen} & PQ\textsubscript{unseen} \\
\hline
CLIP & 41.8 & 53.3 & 33.3 \\
CLIP\textsubscript{\cTwo{}} & \textbf{46.4} & \textbf{58.2} & \textbf{37.6} \\
\bottomrule
\end{tabular}
\vspace{-0.5em}
\label{tab:perfect_masks}
\end{table}

\subsubsection{Training strategy}
Inspired by MAFT+~\cite{jiao2024collaborative}, \cTwo{} adopts a freezing strategy for the CLIP ConvNeXt-L backbone~\cite{radford2021learning, liu2022convnet}, targeting the final MLP and normalisation layers, but discarding their representation consistency (RC) loss. 
To support this choice, we compare \cTwo{} against five variants that employ different strategies to preserve CLIP's embedding alignment. For a controlled comparison, we omit \cOne{} in all variants of this experiment.

We define the variants of the training loss as $\mathcal{L}_c \in {\mathcal{L}, RC, G}$, where $\mathcal{L}$ denotes the base loss introduced in (\ref{eq:loss_class}). The variant $RC = \mathcal{L} + \mathcal{L}\textsubscript{RC}$ incorporates the MAFT+~\cite{jiao2024collaborative} RC loss, while $G = \mathcal{L} + \mathcal{L}_\mathrm{Gram}$ adds a Gram Matrix consistency loss inspired by DINOv3~\cite{simeoni2025dinov3}. 
Furthermore, we define the state of the CLIP~\cite{radford2021learning} image encoder as $s \in {\text{F}, \text{U}}$, where F and U denote a frozen and unfrozen state, respectively. This allows us to denote each model variant as \cTwo{}\textsubscript{$L_c, s$}, with the baseline represented as \cTwo{}\textsubscript{B, F}. The results are shown in ~\Cref{tab:training_ablation}.
\begin{table}[ht!]
\caption{
Comparison of finetuning strategies on ADE20K~\cite{zhou2017scene}.
%\textbf{Ablation study of the CLIP training strategy on ADE20K~\cite{zhou2017scene}.} The table compares the effect of different backbone layer freezing and constraint strategies on panoptic quality (PQ). The RC loss and Gram matrix loss configurations report their best-achieved performance, which ultimately matched the results of our standard frozen configuration.
}
\label{tab:training_ablation}
\centering
\begin{tabular}{l|ccc}
\toprule
Method & PQ & SQ & RQ \\
\midrule
\cTwo{}$_{B, U}$ & 27.2 & 72.4 & 33.1 \\
\cTwo{}$_{RC, U}$ & 27.7 & 73.0 & \textbf{33.7} \\
\cTwo{}$_{G, U}$ & 27.7 & 71.3 & 33.6 \\
\cTwo{}$_{G,F}$ & 27.6 & 73.5 & 33.4 \\
\cTwo{}$_{RC,F}$ & \textbf{27.7} & 71.2 & 33.6 \\
\midrule
\cTwo{} & 27.6 & \textbf{73.6} & 33.4 \\
\bottomrule
\end{tabular}
\end{table}

Training without additional constraints (\cTwo{}\textsubscript{B, U}) results in only a 1.5\% decrease in PQ compared to \cTwo{}, while achieving performance comparable to MAFT+~\cite{jiao2024collaborative} (cf.\ \cref{tab:reclip_results}) with a marginal gain of 0.1 PQ (0.03\%). This robustness is attributed to the stage-one training and the lower learning rate used during the second-stage tuning of CLIP~\cite{radford2021learning}. Aggressively enforcing feature constraints via RC or Gram matrix losses~\cite{jiao2024collaborative, ding2022open} does not have a significant impact on models performance with respect to \cTwo{}. Across all configurations, the best-performing variants achieve results within 1\% of \cTwo{}’s performance. These findings justify freezing the final layers without additional auxiliary losses, providing a favourable trade-off between simplicity and stability while avoiding the complexity and sensitivity introduced by constraint-based training.

\subsubsection{Memory Efficiency}
We further analyse the memory usage of \name{}. Results in~\Cref{tab:memory_usage} show that \name{} requires substantially less memory compared to the state-of-the-art MAFT+~\cite{jiao2024collaborative}.
For a batch size of 1, the average memory usage is reduced by approximately $56\%$ (16.1 GB). 
This is due to the streamlined design of our fine-tuning strategy \cTwo{}, which does not require content-dependent vocabulary embeddings or additional mask-overlap computations.

Considering that the reduction in memory consumption comes with an improvement in $5.5\%$ PQ,  OVRCOAT  brings several practical benefits: it accelerates experimentation, enables use of larger batches (which is particularly important in modern unsupervised learning), and caters to the trend of exploring ever larger models, where memory consumption is a critical limiting factor. Finally, it also enables training on resource-constrained hardware, offering far greater accessibility.

\begin{table}[h]
\centering
\caption{Average training-time memory consumption.% for \name{} and MAFT+ during training.
}
\label{tab:memory_usage}
\begin{tabular}{l|cc}
% \toprule
% Model & MAFT+ & \name{} \\
% \midrule
% Memory (GB) & 27.0 & 14.0 \\
% PQ (ADE20K) & 28.6 & 27.1 \\
% \bottomrule
\toprule
Model & PQ (ADE20K) & Memory [GB/image] \\
\midrule
MAFT+ & 27.1 & 27.0 \\
\name{} & \textbf{28.6} & \textbf{12.5} \\
\bottomrule
\end{tabular} \vspace{-0.5em}
\end{table}

%this is particularly important to accelerate experimentation, enabling the use of larger models or batch sizes, and is essential for training on resource-constrained hardware, offering far greater accessibility. %for practical applications and development cycles of the method adaptations and future extensions.

\begin{figure}[b]
\centering
\includegraphics[width=0.95\linewidth]{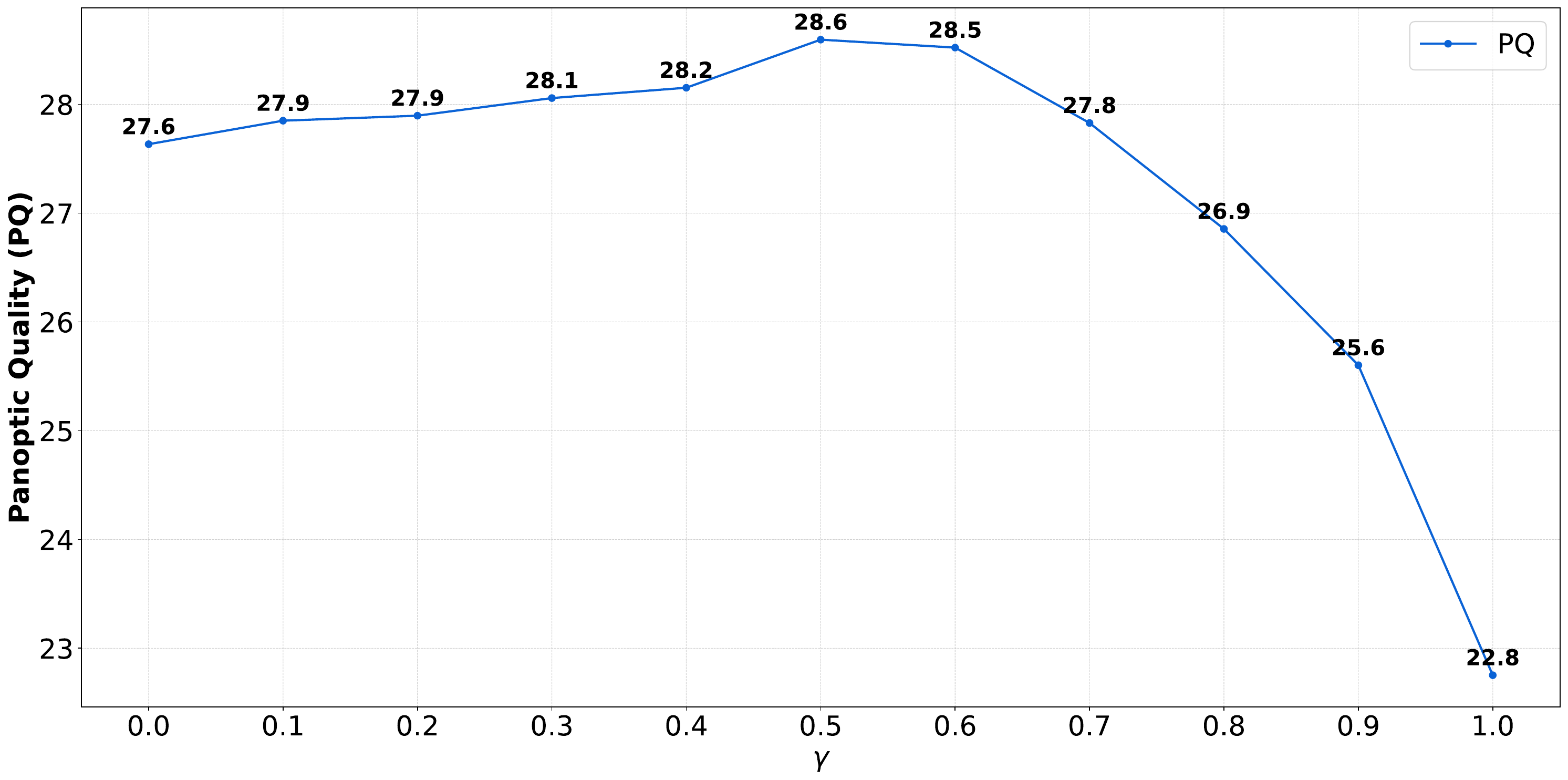}
\caption{
\name{} performance on ADE20K~\cite{zhou2017scene} remains stable over a range of trust factor $\gamma$ values.
%indicates stability.% panoptic quality (PQ). 
%Moderate values improve performance, while excessively high ones lead to a drop in PQ.
}
\label{fig:trust_weight}
\end{figure}

\begin{figure*}[ht!]
\centering
\includegraphics[width=0.95\linewidth]{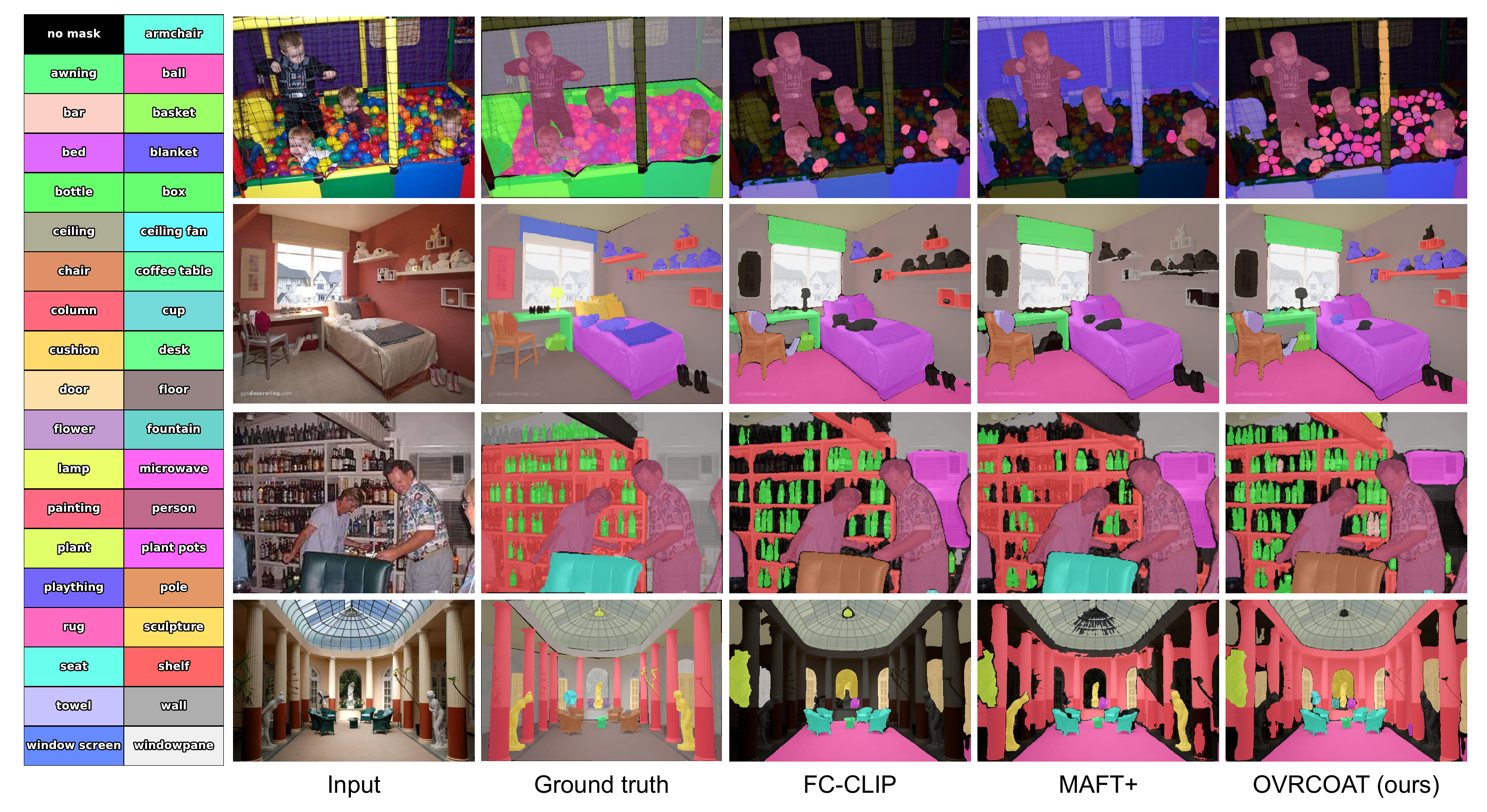}
\caption{
Qualitative analysis. 
%comparison of \name{}, MAFT+~\cite{jiao2024collaborative}, and FC-CLIP~\cite{yu2023convolutions}. 
\name{} consistently recovers more object masks, including regions that MAFT+~\cite{jiao2024collaborative} and FC-CLIP~\cite{yu2023convolutions} misclassify as background, e.g., playthings in the bedroom and bottles in the bar. It also produces more accurate segmentations.
%\name{} consistently recovers a greater number of meaningful object masks and correctly classifies regions that competing models misidentify as background, such as additional playthings in the bedroom and bottles in the bar. It also produces finer, more object-centric segmentations, though this occasionally leads to slightly lower mIoU due to smaller individual masks.
}
\label{fig:qualitative}
\end{figure*}

\subsubsection{CLIP Trust Factor}
As a final validation, we evaluate the sensitivity of \name{} to the CLIP trust weight $\gamma$ used in (\ref{eq:clip_update}). \Cref{fig:trust_weight} reports the PQ values on ADE20K~\cite{zhou2017scene} with respect to different values of $\gamma$. As $\gamma$ increases towards $0.5$, the performance increases to 28.6 PQ, then gradually decreases to 22.8 PQ (a 20\% drop) at $\gamma=1.0$. The graph suggests that \name{} is stable across a range of CLIP trust values, with a broad range of values resulting in top PQ score.

%In \cref{fig:trust_weight}, we evaluated \name{} on ADE20K~\cite{zhou2017scene} for different values of the trust weight for our CLIP-based void update. Performance improves to 28.6 as the weight increases to 0.5, after which it gradually decreases, reaching 22.8 PQ at a weight of 1.0.

\subsection{Qualitative analysis}

%Having analysed how individual components contribute to the quantitative gains in panoptic quality, we next examine their qualitative impact. 
\Cref{fig:qualitative} provides qualitative comparison of \name{} with FC-CLIP~\cite{yu2023convolutions} and MAFT+~\cite{jiao2024collaborative}. We observe that \name{}
consistently identifies more meaningful object masks than the state-of-the-art, supporting the generalization improvements reflected in previous experiments. 
In the \textit{bedroom} and \textit{children} scenes, \name{} successfully recovers additional playthings and balls that competing models misclassify as background. In the \textit{museum} example, \name{} detects finer architectural details such as doors and pillars. %, though MAFT+ achieves slightly better performance on statues. 
Moreover, \name{} produces finer, object-centric segmentations. For instance, it separates multiple bottles in the \textit{bar} scene instead of a larger shelf mask as in MAFT+~\cite{jiao2024collaborative}. 
This behaviour highlights a strengthened general object understanding, particularly benefiting the detection of smaller and unseen objects.

% may explain our lower mIoU, as the larger mask has a better mIoU than multiple smaller ones. 
%the latter is unstable for small objects.
%slightly reduce mIoU, as smaller, individual masks overlap less with merged ground-truth regions.

%% file: sec/5_conclusion.tex
\section{Conclusion}
% We presented \name{}, a simple yet effective framework that establishes a new state-of-the-art in open-vocabulary panoptic segmentation. The core of our approach lies in combining a carefully fine-tuned CLIP backbone with a novel CLIP-guided void probability update. This design successfully addresses the major challenge of mask selection bias by reducing misclassifications and preserving semantic alignment. Through extensive ablations, we validated the simplicity of our fine-tuning strategy, showing that complex feature constraints, such as representation consistency or Gramme-matrix losses, do not provide additional performance benefits.

% Experiments demonstrate that \name{} achieves consistent performance gains by enhancing the segmentation of unseen classes while simultaneously maintaining competitive accuracy on seen classes. Furthermore, \name{} offers reduced complexity and memory usage compared to prior approaches. Overall, \name{} provides a robust, efficient, and versatile framework that advances open-vocabulary panoptic segmentation across diverse domains.

We addressed two pressing obstacles in open-vocabulary panoptic segmentation: the mask-selection bias induced by mask transformer training on a closed vocabulary dataset, and reduced accuracy of region-level understanding in vision–language models trained for global classification. We introduced OVRCOAT, a minimal and modular framework that tackles both: \cOne~adjusts background/foreground probabilities using CLIP to retain object masks for unseen categories, while \cTwo~refines mask-to-text alignment to improve recognition of both seen and unseen classes.

OVRCOAT’s design brings practical benefits: clean integration into standard Mask2Former pipelines and substantially lower memory demands than recent fine-tuning approaches, while remaining robust across datasets. Empirically, OVRCOAT consistently surpasses prior open-vocabulary panoptic methods on challenging out-of-vocabulary benchmarks, and ablations verify the individual components. \cOne~reduces selection bias and boosts recall of novel objects, \cTwo~strengthens regional alignment, and their combination yields the most reliable panoptic improvements. Analyses with oracle masks confirm the alignment gains, probability-fusion studies favor combining CLIP with mask-transformer scores, and training-strategy experiments show that strong results are attainable without complex representation-consistency losses.

Looking forward, we see OVRCOAT as a foundation for broader open-world perception, extending to video panoptics with temporal consistency, coupling with promptable or interactive segmentation.
As stronger VLMs and data engines emerge, OVRCOAT’s modularity should make such upgrades straightforward, further closing the gap between closed-set training and open-world deployment.

%% file: sec/6_acknowledgements.tex
\section*{Acknowledgements}

This work was supported in part by research programs and projects P2-0214, J2-60054 and SLAIF grant no. 101254461 and by ARNES/SLING supercomputer clusters. Josip Šarić was funded from the European Union‘s Horizon Europe research and innovation program under the Marie Sklodowska Curie COFUND Postdoctoral Programme grant agreement No.101081355- SMASH, the Republic of Slovenia and the European Union from the European Regional Development Fund.
Disclosure: Co-funded by the European Union. Views and opinions expressed are however those of the author(s) only and do not necessarily reflect those of the European Union or European Research Executive Agency. Neither the European Union nor the granting authority can be held responsible for them.

%% file: sec/7_supplementary.tex
%\title{OVRCOAT: Mitigating Objectness Bias and Image-to-Text Alignment for Open-Vocabulary Panoptics}
\clearpage
\setcounter{page}{1}        
\setcounter{section}{0}   
\renewcommand{\thesection}{\Alph{section}} 
\maketitlesupplementary
\begin{figure}[b!]
    \centering
    \includegraphics[width=2.1\linewidth]{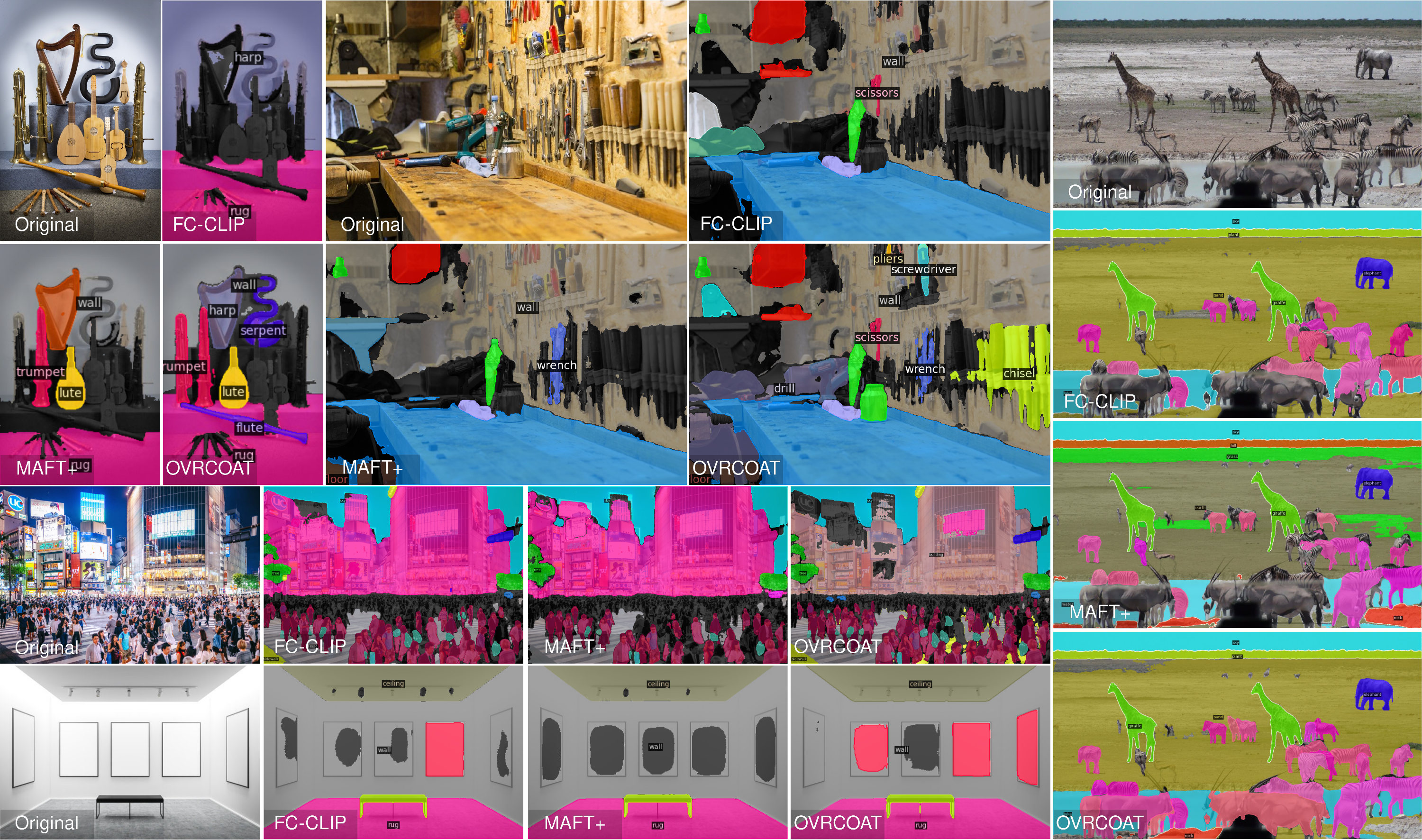}
    
    \label{fig:out-of-vocabulary}
    \vspace{5pt} % Adjust spacing between image and caption
    
    % Force the caption to be full-page width and centered
    \makebox[\textwidth][c]{
        \begin{minipage}{\textwidth}
            \centering
            \captionof{figure}{Qualitative comparison on unconventional scenes using an extended ADE20K vocabulary.}
            \label{fig:out-of-vocabulary}
        \end{minipage}
    }
\end{figure}

\section{Per-Image Inference Latency}
We measured the per-image inference time on the ADE20k validation set to assess computational efficiency.  
All three methods run at comparable speeds: 0.31\,s for FC-CLIP~\cite{yu2023convolutions}, 0.30\,s for MAFT+~\cite{jiao2024collaborative}, and 0.32\,s for \name{}.  
The maximum difference of 0.02\,s indicates that the proposed method delivers competitive performance without introducing significant computational overhead.

\section{Extended Qualitative Analysis}
We provide five example images from the internet~\cite{Booth2021,renoWorkshopImage2024,tips2024_unsichtbarekunst, shibuya2019_cheapoguides,youtube_thumbnail_C-4x4QpmtMs} to highlight model performance. In three cases, the ADE20K~\cite{zhou2017scene} vocabulary is extended with musical instruments, workshop tools, or safari animals (Fig.~\ref{fig:out-of-vocabulary}). \name{} outperforms both FC-CLIP~\cite{yu2023convolutions} and MAFT+~\cite{jiao2024collaborative}, leveraging COAT to recover masks and OVR to classify them more accurately. Nevertheless, certain instruments, such as flutes and the baroque guitar, are not detected, likely due to their unconventional appearance. Tool masks are also suboptimal, with oversegmentation (e.g., the drill) and undersegmentation (e.g., the wrench and chisel) in the unseen setting. All models miss distant animals on the safari scene, likely due to their small size. These examples demonstrate the superior generalisation capabilities of \name{}, while exposing common limitations.

The museum scene evaluates robustness to visually ambiguous content: blank paintings matching the wall colour. \name{} detects the most paintings, while MAFT+~\cite{jiao2024collaborative} fails to detect any. One of the paintings is incorrectly included in the wall segment by \name{}, highlighting remaining segmentation challenges.

Finally, the Shibuya scene illustrates a real-world scenario where all models perform similarly. \name{} correctly classifies buildings and segments more civilians, but fails to capture five signboards in ambiguous regions. This underscores that such areas remain a significant challenge for open-vocabulary models and represent a key area for future investigation into segmentation robustness.

\clearpage